# What Radio Waves Tell Us about Sleep

Hao He[1*], Chao Li[1*], Wolfgang Ganglberger[2,3,5], Kaileigh Gallagher[5], Rumen Hristov[6], Michail Ouroutzoglou[1], Haoqi Sun[2,5], Jimeng Sun[4], Brandon Westover[2,3,5], Dina Katabi[1,6]

[1]Department of Electrical Engineering and Computer Science, Massachusetts Institute of Technology, Cambridge, MA, USA

[2]McCance Center for Brain Health, Massachusetts General Hospital, Boston, MA, USA

[3]Division of Sleep Medicine, Harvard Medical School, Boston, MA, USA

[4]Computer Science Department, University of Illinois Urbana-Champaign, Urbana, IL, USA

[5]Department of Neurology, Beth Israel Deaconess Medical Center, Boston, MA, USA

[6]Emerald Innovations Inc., Cambridge, MA 02142, USA

* These authors contributed equally to this work.

Corresponding author. Hao He, Department of Electrical Engineering and Computer Science, Massachusetts Institute of Technology, Cambridge, MA 02139, USA. Email haohe@mit.edu; Corresponding author. Chao Li, Department of Electrical Engineering and Computer Science, Massachusetts Institute of Technology, Cambridge, MA 02139, USA. Email chaoli@mit.edu; Corresponding author. Dina Katabi, Department of Electrical Engineering and Computer Science, Massachusetts Institute of Technology, Cambridge, MA 02139, USA. Email dk@mit.edu;

## Abstract

The ability to assess sleep at home, capture sleep stages, and detect the occurrence of apnea (without on-body sensors) simply by analyzing the radio waves bouncing off people's bodies while they sleep is quite powerful. Such a capability would allow for longitudinal data collection in patients' homes, informing our understanding of sleep and its interaction with various diseases and their therapeutic responses, both in clinical trials and routine care. In this article, we develop an advanced machine learning algorithm for passively monitoring sleep and nocturnal breathing from radio waves reflected off people while asleep. Validation results in comparison with the gold standard (i.e., polysomnography) (n=880) demonstrate that the model captures the sleep hypnogram (with an accuracy of 80.5% for 30-second epochs categorized into Wake, Light Sleep, Deep Sleep, or REM), detects sleep apnea (AUROC = 0.89), and measures the patient's Apnea-Hypopnea Index (ICC=0.90; 95% CI = [0.88, 0.91]). Notably, the model exhibits equitable performance across race, sex, and age. Moreover, the model uncovers informative interactions between sleep stages and a range of diseases including neurological, psychiatric, cardiovascular, and immunological disorders. These findings not only hold promise for clinical practice and interventional trials but also underscore the significance of sleep as a fundamental component in understanding and managing various diseases.

## Statement of Significance

Our research presents an important advancement in sleep medicine, achieving accurate contactless sleep staging and detection of respiratory events (compared to PSG), through contactless radio wave measurements. Demonstrating robust and consistent performance across diverse age groups, genders, races, and health conditions in a large population of sleep clinic patients, our methodology facilitates daily, passive, at-home sleep assessment. This advancement holds promise for enhancing clinical care and clinical trials by providing longitudinal, at-home, cost-effective alternative to traditional in-laboratory sleep studies. Our findings have the potential to transform the long-term monitoring of sleep-related health issues, contributing to a more personalized, accessible, and data-driven approach to healthcare.

# Introduction

Sleep monitoring is proven pivotal in understanding diseases and drug response. Sleep is implicated in almost all neurological diseases and inflammatory conditions [1, 2] Sleep alterations in patients with depression include reduced REM latency and increased REM density [3, 4], and popular depression drugs, such as SSRI and SNRI, counter these effects increasing REM latency and reducing REM density [5]. Slow waves during deep sleep correlate with high activity of the glymphatic system and reduced slow waves have been associated with a greater β-amyloid burden, a hallmark of Alzheimer's disease [6]. Past work has also shown that disrupted sleep can aggravate inflammatory pathways present in conditions like rheumatoid arthritis and inflammatory bowel disease [7]. These examples emphasize the importance of sleep monitoring and assessment for both clinical trials and patient care.

The gold standard for sleep assessment is polysomnography (PSG) [8], which requires patients to sleep in the clinic with a plethora of electrodes and sensors on their bodies (Figure 1A). A key output of a PSG is the sleep hypnogram, which reports for each 30-second epoch the patient's sleep stage (Figure 1B). PSG studies are also used to assess irregular breathing patterns like apnea (cessation of breathing for more than 10 seconds) and hypopnea (shallow breathing for more than 10 seconds) (Figure 1B). However, PSG is costly and incurs a high patient burden [9, 10]. Furthermore, sleeping in an unfamiliar environment with many on-body sensors (as shown in Fig 1A) can bias the assessment [11, 12]. These challenges are exacerbated in longitudinal studies where changes in sleep and nocturnal breathing are tracked over months or years. As a result, both clinical trials and clinical care rarely use PSG but rather rely on subjective and insensitive self-reports of sleep quality (e.g., via Sleep NRS) by patients [13, 14].

Both clinical trials and patient care would benefit from an objective, sensitive, and low burden solution for in-home assessment of sleep hypnogram and nocturnal breathing. Indeed, the last decade has witnessed increasing use of wearable devices [15-18] and in-mattress sensors for measuring sleep at home in clinical trials [19-21]. However, wearable solutions typically require multiple on-body sensors to concurrently track sleep and respiratory signals [16-18], which imposes a relatively high burden in longitudinal trials, and can be cumbersome to wear during sleep. Mattress sensors can similarly affect a person's sleep and have relatively low accuracy [19-21].

Recently, an alternative completely passive approach was proposed (by our lab and others [22-26]: a radio device similar to a WiFi router is placed in the bedroom of the patient. It transmits very low power radio waves (1000 times lower power than home WiFi) and analyzes their reflections using neural networks to infer the sleep hypnogram and the respiration signal (Figure 1D). This approach is referred to as RF-based sleep monitoring (as it leverages Radio Frequency (RF) signals) and is as safe as having a WiFi router at home.

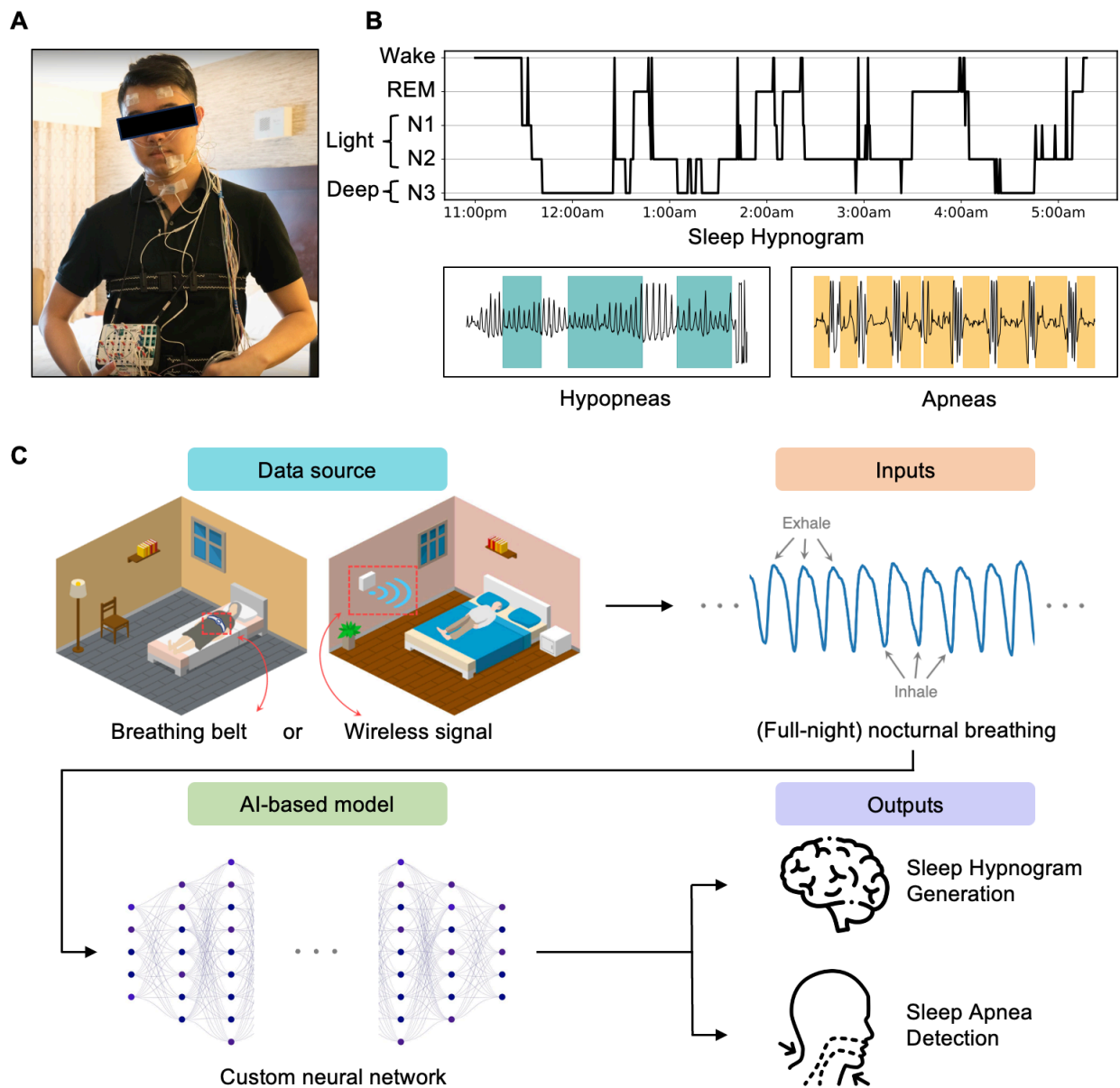

**Figure 1. Illustration of traditional and RF-based sleep assessment. (A)** PSG sensors during a sleep study. **(B)** Illustration of the output of sleep studies showing an example sleep hypnogram with 5 distinct sleep stages**,** and examples of respiratory signals with hypopnea events marked in cyan, and apnea events marked in yellow. **(D)** Illustration of our RF-based sleep and apnea monitoring system. Our sensor operates like a low power radar. It transmits wireless signals that are 1/1000 the power of a home Wi-Fi router and collects their reflections from nearby people. The breathing signal is extracted from such reflections, and the sleep hypnogram and respiratory events are then predicted from each night of the nocturnal breathing signal.

RF-based sleep monitoring promises to benefit both clinical trials and clinical care by allowing longitudinal sleep and apnea monitoring of patients in their own homes for months or even years, at zero overhead to patients. Furthermore, it concurrently measures sleep metrics and nocturnal breathing signals. Yet, achieving this potential depends on whether RF-based monitoring works well for patients across diverse demographics and different health statuses.  Past work on RF-based sleep monitoring focuses on wellness as opposed to clinical use and hence is validated on a small number of healthy young users, with no consideration to demographics or pre-existing diseases. Hence, the validation of RF-based sleep monitoring for clinical purposes remains an unmet need.

In this article, we explore the feasibility and accuracy of using RF-based monitoring of sleep and sleep apnea for clinical purposes. We focus on important clinical questions such as: does RF-based monitoring work well for the types of patients evaluated in sleep clinics? Is the performance equitable for different races, sexes, and age groups? Can we trust it for people who have comorbidities? Is the performance impacted by certain comorbidities more than others? And finally, is RF-based monitoring sensitive enough to capture the interplay between sleep and various diseases? Answering these questions is essential for understanding the clinical value of RF-based sleep monitoring.

To investigate the above questions, we developed a new advanced machine learning algorithm for monitoring sleep and nocturnal breathing from radio waves reflected off people while asleep. The model first extracts nocturnal breathing from radio signals; then breathing is analyzed by two neural networks- the first of which outputs the sleep hypnogram and the second outputs apnea/hypopnea events (Figure 1C). By using breathing signals as an intermediate representation, the model can be trained and tested, not only on data from radio waves but also on nocturnal breathing from repositories of public sleep datasets. Furthermore, the model is explicitly designed and trained to ensure equitable performance on different demographics. A detailed description of the model is provided in the Methods section.

This new AI model allows us to positively answer all of the aforementioned questions. Specifically, the model was evaluated on a large dataset comprising 12,672 individuals, using data extracted from radio waves (n=880) and breathing belts commonly used in sleep studies (n = 12,672). Our study reveals the following findings:

- Sleep hypnograms generated by the AI model solely from radio waves have an average accuracy of 80.5% and a standard deviation of 8.1%. Accurate sleep metrics are extracted from radio waves including sleep efficiency ($r=0.88$, $p<0.001$), total sleep time ($r=0.92$, $p<0.001$), sleep latency ($r=0.86$, $p<0.001$), and wake after sleep onset (WASO) ($r=0.87$, $p<0.001$).
- The same model can also generate a sleep hypnogram using data from a breathing belt with an average accuracy of 81.8% and a standard deviation of 8.4%. This accuracy persists on external datasets unseen during training.
- The model demonstrates equitable performance, where the performance gap stays small across races (1.47%), sexes (0.60%), and age groups (1.65%).
- The model can also detect sleep apnea from radio signals (AUC = 0.89) and assess the patient's Apnea-Hypopnea Index ($r=0.82$, $p<0.001$).
- The model continues to deliver high performance even in the presence of existing diseases, including neurological, psychiatric, cardiovascular, and immunological disorders. Notably, though the model relies on breathing signals as an intermediate representation from which it generates sleep stages, the model's performance is not sensitive to whether the person has pulmonary diseases (e.g., COPD, bronchitis, or emphysema).
- Interestingly, the model also reveals differences in sleep architecture in depression, hypertension, rheumatoid arthritis, and diabetes, underscoring the interplay between sleep and such conditions ($p < 0.05$).

## Methods

### Datasets

We use a large and diverse population created by pooling several datasets from different sources, including Massachusetts General Hospital (MGH) sleep lab, University of Massachusetts (UMASS) sleep lab, and public sleep datasets from the National Sleep Research Resource including the Sleep Heart Health Study (SHHS), Wisconsin Sleep Cohort (WSC), Multi-Ethnic Study of Atherosclerosis (MESA), and MrOS Sleep Study (MROS). The combined dataset contains 17,676 nights with over 161,091 hours of nocturnal breathing signals from 12,672 individuals aged 66.7 (standard deviation of 11.9), of whom 39.9% are females and 81.8% are white participants. Table 1 summarizes the datasets.

| | Cohort/Dataset Type | Dataset Usage | Breathing Signal Source | No. of Participants | No. of Nights | Total Hours | Sex (%) | Age [years] mean (SD) | Race (%) | | | |
|---|---|---|---|---|---|---|---|---|---|---|---|---|
| | | | | | | | Males | | Asian | Black | White | Others |
| **MGH** | MGH Sleep Lab Population | Internal | Wireless & Belts | 849 | 849 | 6412 | 56.9 | 54.1 (16.7) | 4.3 | 7.9 | 78.9 | 8.9 |
| **UMASS** | UMASS Sleep Lab Population | Internal | Wireless & Belts | 31 | 31 | 273 | - | - | - | - | - | - |
| **SHHS** (visit 1 & 2) | Sleep Disorders and Cardiovascular Outcomes | Internal | Breathing Belts | 5777 | 8381 | 72043 | 47.1 | 64.5 (11.1) | 0.0 | 8.2 | 85.4 | 6.4 |
| **WSC** | Longitudinal Study of Sleep Disorders | Internal | Breathing Belts | 1119 | 2514 | 20038 | 54.1 | 59.7 (8.5) | 0.9 | 1.4 | 95.7 | 2.0 |
| **MESA** | Sleep Disorders and Cardiovascular Disease | External | Breathing Belts | 2021 | 2021 | 21221 | 46.4 | 69.3 (9.1) | 12.0 | 27.9 | 36.3 | 23.8 |
| **MROS** (visit 1 & 2) | Sleep Disorders, Fractures and Vascular Disease | External | Breathing Belts | 2875 | 3880 | 41104 | 100 | 77.6 (5.6) | 3.5 | 3.5 | 90.0 | 3.0 |

Dashes, unavailable data.
For dataset usage, 'internal' means the dataset is used for training and cross validation; 'external' means the dataset is held out for testing purposes only.

**Table 1. Demographics of the datasets in the study.**

The datasets can be categorized into two groups. The first group contains nocturnal breathing collected in a contactless manner using wireless radio signals. The radio sensor was deployed in the sleep lab; it analyzed the radio reflections from the environment to extract the person's breathing signal, which was further processed to infer the person's hypnogram and apnea events. This group of datasets allows us to confirm that our model can operate passively using only the radio signals reflected off a person's body during their sleep. The second group contains nocturnal breathing signals collected using (abdominal and thoracic) breathing belts during PSG sleep studies. This group of datasets allows us to confirm that the model works well independent of whether the breathing signals were captured using radio waves or a wearable device. It also allows us to further confirm the accuracy of the model for large datasets collected in different studies and with diverse populations. (Further description of the datasets and the corresponding IRBs is provided in the Supplementary Material.)

**Deep Learning Models**

As mentioned earlier, our models operate on breathing signals extracted from either radio reflections or wearable breathing belts. In this paper, we focus on the machine learning models and their clinical validations and leverage past work for extracting accurate breathing signals from radio reflections. As in past work [22, 27], the wireless signals were captured using a standard multi-antenna frequency-modulated continuous waves (FMCW) radio. We extract the person's breathing from the wireless signal using the method developed by Yue et al., which has been shown to achieve high accuracy [22]. (More information about the radio device can be found in the Supplementary Material.)

Our sleep staging model is built upon a deep neural network architecture. It takes as input an entire night's worth of breathing signals and generates a sleep hypnogram. This hypnogram represents a time series of sleep stages, with each stage assigned every 30 seconds. The model's architecture comprises 11 convolutional residual blocks, which are responsible for identifying local patterns within the breathing signals. Additionally, it incorporates a long short-term memory (LSTM) layer and attention layer, facilitating the aggregation of information across the entire sequence of signals. Finally, the model employs three convolutional layers at the end to predict sleep stages based on the extracted latent features. (Refer to Supplementary S1 for model architecture.)

To improve the accuracy of our sleep staging model, we leverage the electroencephalogram (EEG) signals from sleep datasets using a technique called knowledge distillation [28, 29]. Knowledge distillation involves two models: student and teacher. In this work, our student model is the model that takes in breathing signals to predict sleep stages, and our teacher model is the model that takes in EEG signals to predict sleep stages. During training, the features learned by the teacher are distilled into the student model in two ways. First, we applied an L2 consistency loss between the features from the student and teacher models to encourage the student model to learn sleep staging features associated with EEG. Next, we applied a cross-entropy loss between the student and teacher model's outputs. Most importantly, the teacher is used only during training. During inference, only the student model is used and hence there is no need for EEG signals.

To improve the sleep staging model's performance across different subpopulations, we employ a state-of-the-art technique called classifier re-training (CRT) [30]. In this technique, the classifier is designed as a combination of an encoder and a prediction head. The classifier is trained in two stages. In the first stage, the model is trained without weighting or re-sampling. Then, the encoder is fixed, and the rest of the model, i.e., the prediction head, is trained. During this retraining phase, data samples are weighted differently to account for data imbalance. In our study, we observed a notable imbalance among racial groups, with a substantially smaller number of Black participants compared to White and other races. To address this, we retrained the last three layers of our model (i.e., the projection head) using a resampled training set where Black participants are given eight times more weight than non-Black participants.

In the Supplementary Material, we present ablation results that demonstrate the benefits of knowledge distillation (Figure S2(A)) and classifier retraining (Figure S2(B) in improving the sleep model's accuracy and equitable performance.

Our apnea model is also a deep neural network that takes as input a full night of breathing signals and predicts the likelihood of an apnea/hypopnea happening at each moment. The model comprises 7 convolutional layers with residual connections. The output of the neural network is post-processed to eliminate apnea/hypopnea events shorter than 10 seconds, and merge events that are within 1 second because they are part of the same apnea/hypopnea event.

**Training and evaluating protocols**
We train a single model that can operate on breathing signals from radio waves or breathing belts. The model is trained using data from MGH, SHHS, WSC, and MESA, while the UMASS and MROS datasets are external datasets that were only used for testing. During training, we randomly split the patients into four folds so that for datasets with multiple visits, all the visits from the same patient will either be in the training set or the test set. We trained the model on three folds and tested on the left-out testing fold. This procedure was repeated four times so that the four testing folds covered the whole dataset. All the performance metrics and statistical analysis are based on the pooled predictions across the four testing folds. (Additional information about training such as hyper-parameters is provided in the Supplementary Material.)

All neural network models in this paper are implemented in Python using PyTorch version 2.2.1.

**Statistical analysis and evaluation metrics**
We evaluated our models using standard metrics. In particular, the accuracy of the sleep staging classifier is computed by comparing the prediction for every 30-second epoch with the ground truth stage, and reporting the average and standard deviation across subjects in each dataset. We used Cohen's Kappa [31] to assess agreement between predictions made using different breathing channels (thorax belt, abdominal belt, and wireless reflections). We also used the Pearson correlation and the corresponding p-value (derived from a one-sided Wald Test with a t-distribution of the test statistic) [32]. A p-value smaller than 0.05 is considered statistically significant.

To evaluate the effectiveness of the apnea model, we calculated the intraclass correlation coefficient (ICC) [33] between the predicted Apnea-Hypopnea Index (AHI) and the ground truth, along with its corresponding 95% confidence interval. We also assessed the model's ability to distinguish between apnea severity categories (Normal, Mild, Moderate, Severe) using the one-sided Mann-Whitney U rank test [34].

To investigate whether our predicted sleep metrics capture changes in sleep architecture associated with diseases and drug effects, we used a linear regression model [35] as follows:
$$\text{Sleep Metric} = \beta_0 + \beta_1 \cdot \text{Variable} + \beta_2 \cdot \text{Age} + \beta_3 \cdot \text{Sex},$$

where Sex is 0 for females and 1 for males, Variable is 1 if the participants have the disease or take the drug, and 0 otherwise. We examined the p-value and the magnitude of the coefficient of the Variable ($\beta_1$) to determine its effect on sleep metric. For all the linear models used, we verified that there is no strong collinearity between the covariates by checking that the Variance Inflation Factor (VIF) for each covariate is below 10.

The statistical analysis was conducted using Python version 3.8 (Python Software Foundation).

## Results

**Evaluation of Predicted Sleep Hypnogram and Sleep Metrics**

In this section, we evaluated its accuracy in predicting the person's sleep hypnogram. We differentiate between the accuracy of extracting the sleep hypnogram directly from radio signals (where breathing is extracted as an intermediate stage) and extracting it from breathing signals collected using breathing belts. As shown in Figure 2A, the model predicts the sleep hypnogram from radio signals with an accuracy of 80.4%±8.2% for the MGH dataset (n=849) and 84.1%±4.2% for the UMass dataset (n=31), (for 30-second epochs categorized into Wake, Light Sleep, Deep Sleep, or REM). For comparison, the model predicts the sleep hypnogram from the abdominal breathing belt in the MGH dataset (n=849) with an accuracy of 81%±8.0%, (for 30-second epochs categorized into Wake, Light Sleep, Deep Sleep, or REM). Figure 2A further lists the accuracy of all datasets. The differences in accuracy can be attributed to differences in the quality of the sensors in older datasets (e.g. SHHS visit 1 collected from 1995 to 1998), which exhibit more noisy measurements and hence lower accuracy. Interestingly, the accuracy of the AI model is on par with the agreement between different sleep technicians who label the same sleep data (which is about 83% [36, 37]). Additionally, we tested our sleep staging model on datasets that were not used in training the model. On these datasets (MROS and UMass) our model similarly showed high performance, demonstrating its ability to generalize to external datasets.

The sleep hypnogram that the model generates from the radio signals is in high agreement with the hypnogram it generates from the breathing collected by the thorax or abdominal breathing belts worn by the participant on the same night. Specifically, Figure 2B shows Cohen's Kappa between predicted hypnograms (4 stages) from two different channels. For any pair of two channels, Cohen's Kappa is around 0.8 indicating a high level of agreement.

We further analyze the model's performance for different sleep stages. Figure 2C shows the confusion matrix for the AI model for classifying every 30-second epoch into one of the following sleep stages: Wake, Light Sleep, Deep Sleep, and REM. It shows that all sleep stages have a relatively high accuracy, with the most common misclassification involving identifying deep sleep as light sleep. This is not surprising since sleep technicians typically differentiate light sleep from deep sleep using subjective thresholds in the delta power band, and these thresholds can vary from one labeler to another [38]. Figure 2B further differentiates light sleep into two stages N1 and N2, the former is an unstable transition state, and the latter is the more stable light sleep. The figure shows that the model can differentiate these two modes for light sleep but faces challenges in predicting the N1 stage due to its brief and transitional nature. This difficulty in predicting N1 is acknowledged in previous studies [39-42], as even sleep technicians struggle with accurately labeling this stage, resulting in a low interrater agreement of 0.24 [43].

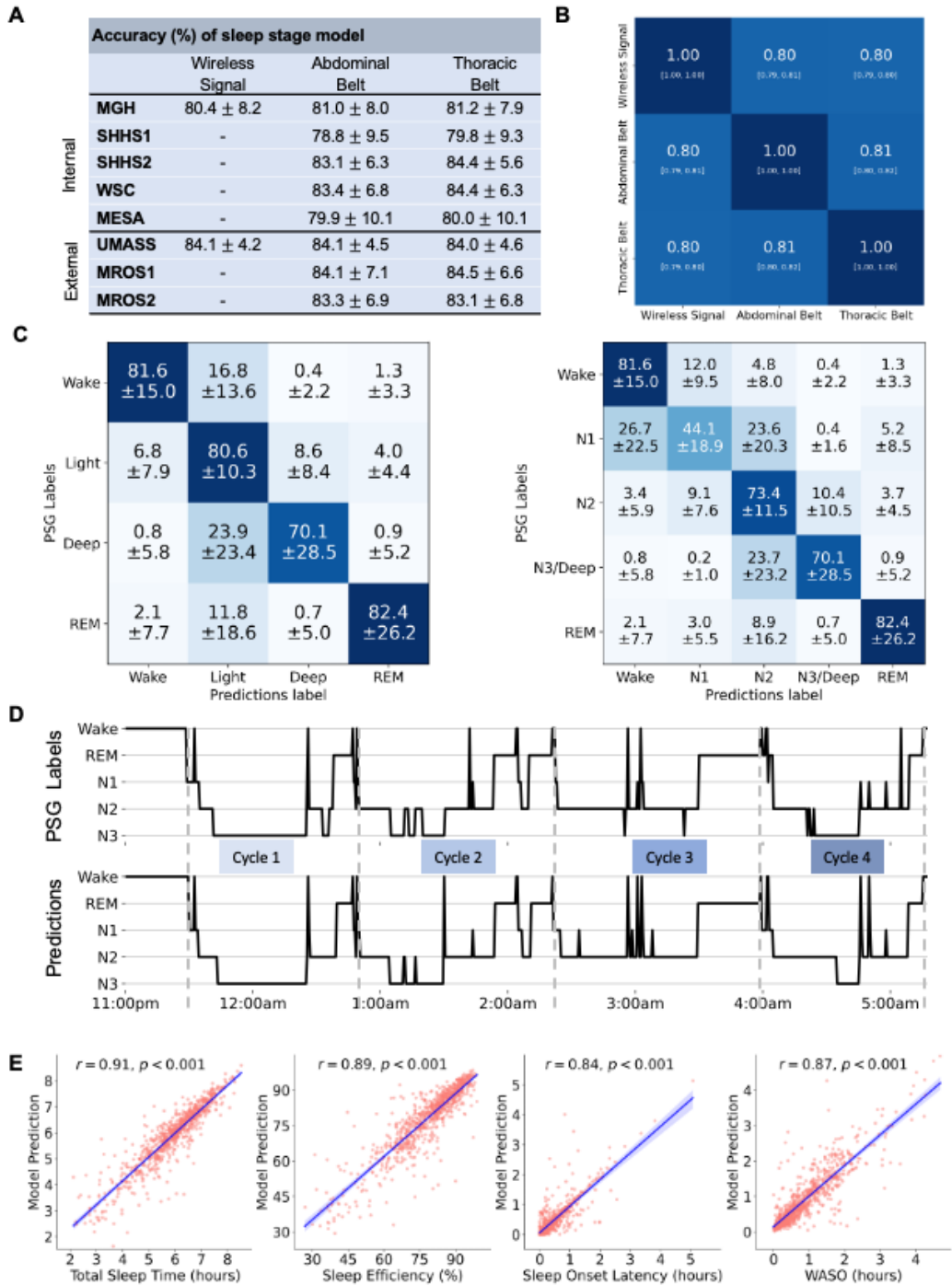

**Figure 2. Performance of our sleep staging model. (A)** Accuracy of staging model across breathing channels and datasets. **(B)** Agreement of model's prediction from different breathing channels, evaluated by Cohen's Kappa. Each entry shows the mean and 95% confidence interval. **(C)** Confusion matrices of predicted sleep stages from radio waves in the MGH dataset. Left is 4 class staging; Right is 5 class staging. **(D)** Example of a ground truth hypnogram from PSG data and the corresponding hypnogram output by our model based on radio signals. The 4 sleep cycles during the night are separated by the dashed lines. **(E)** Scatter plots of predicted sleep metrics from radio signals versus the ground truth from PSG data. Metrics are Total Sleep Time (TST), Sleep Efficiency (SE), Sleep Onset Latency (SOL), and Wake After Sleep Onset (WASO). Pearson correlations and p-values from the two-tailed Wald Test with t-distribution are reported.

Figure 2D shows an example of the predicted sleep hypnogram and the corresponding PSG-labeled sleep hypnogram. They highlight the model's ability to capture the overarching sleep architecture and distinct sleep cycles of the patient.

Finally, we consider changes in sleep quality assessed in terms of four sleep metrics: (1) total sleep time; (2) sleep efficiency (i.e., fraction of the sleep time in which the person is asleep); (3) sleep latency (i.e., time spent awake before sleep onset); (4) WASO (i.e., amount of wake time after sleep onset). Figure 2E shows that the AI model extracts accurate sleep metrics from radio waves including sleep efficiency (r=0.89, p<0.001), total sleep time (r=0.91, p<0.001), sleep latency (r=0.84, p<0.001), and WASO (r=0.87, p<0.001). The figure is for the MGH dataset. The results from other datasets are similar regardless of whether the input to the model is collected via radio signals or breathing belts (as shown in Figure S4 in the Supplementary Material).

**Equitable Performance Across Demographics**

Machine learning models often perform poorly on minority groups since they are under-represented in the training data [46]. To address this issue, our machine learning model uses "Classifier Retraining" [30], a state-of-the-art design that addresses data imbalance and improves minority performance. In this section, we show that the model performs well on minority groups and the performance (or accuracy) gap for different demographics is negligible.

The performance gap between the two groups is defined as the absolute difference between the model's average accuracy on the two groups. Specifically, for sex, we examined the performance gap between males and females. For race, we categorized individuals into four groups: Asian, Black, White, and Others, and calculated the performance gaps between each pair of groups. For age, we defined three categories: Young (under 40 years), Middle-aged (40 to 59 years), and Senior (60 years and over), and computed the average performance gap between them.

Table 3A reports the mean and standard deviation of the performance gap for different sexes, ages, and races, in the studied datasets. The table shows that the performance gap is negligible (<1.7%), demonstrating that the model maintains good performance independent of the person's sex, race, and age. Among the three demographics, the performance across different sexes has the smallest gap (0.60%) which is attributed to the fact that the training datasets have balanced numbers of males and females. We also note that the gap between different age groups is the largest in datasets that span a wide range of ages (i.e., SHHS1, WSC), which is expected.

To further check that our model achieves equitable performance across demographics when using wireless signals, we zoom in on performance for the MGH wireless dataset. Table 3B lists the accuracy and Cohen's kappa for 4-class sleep staging in each demographic group. Notably, the disparity between the highest and lowest performing groups is marginal, at 2.9% for accuracy and 0.061 for Cohen's kappa. We further conduct hypothesis testing to show that such gaps are not statistically significant. We set the null hypothesis to be that the two groups have the same mean accuracy. The resulting p-value hence indicates the significance of the accuracy gap. Across the different sexes, statistical testing reveals negligible differences in mean accuracy (p = 0.90). For age, the result shows young and middle-aged groups have close accuracy (p = 0.10). However, the performance of senior participants is observed to be lower. This is explainable as the EEG signals (used in labeling sleep stages in a PSG study) among older participants are less reliable, leading to more uncertainty in sleep stage labels [45]. For race, the p values show that the gap is not significant (p = 0.12 for White versus Black and p = 0.08 for White versus Asians).

**A**

Performance gap of the sleep stage model for different demographics

| Dataset | Accuracy gap (%) (mean ± std) | | |
|---|---|---|---|
| | for Sex | for Age | for Race |
| **MGH** | 0.22 | 1.54 ± 0.74 | 1.21 ± 0.59 |
| **SHHS1** | 0.06 | 2.19 ± 0.83 | 1.72 ± 0.61 |
| **SHHS2** | 0.48 | 0.51 ± 0.00 | 1.23 ± 0.46 |
| **WSC** | 1.41 | 2.24 ± 0.85 | 1.05 ± 0.59 |
| **MESA** | 0.80 | 1.79 ± 0.00 | 1.86 ± 1.06 |
| **MROS1** | NA | NA | 1.85 ± 1.04 |
| **MROS2** | NA | NA | 1.34 ± 0.71 |
| **Average** | 0.60 | 1.65 | 1.47 |

NA means "not applicable". All MROS participants are males with age >= 60.
Sex groups: Male, Female
Age groups: Young (age < 40), Middle (40<=age<60), Senior (age >= 60)
Race groups: Asian, Black, White, Others

**B**

Performance across demographics, on MGH dataset, using wireless signals as inputs

| Demographic | No. of samples | Accuracy (%) | Cohen's Kappa |
|---|---|---|---|
| All | 849 | 80.4 ± 8.2 | 0.681 ± 0.140 |
| *Gender* | | | |
| Male | 480 | 80.4 ± 7.9 | 0.679 ± 0.138 |
| Female | 363 | 80.2 ± 8.6 | 0.681 ± 0.144 |
| *Age* | | | |
| Young (<40) | 202 | 81.5 ± 8.7 | 0.709 ± 0.138 |
| Middle [40,60] | 287 | 81.0 ± 8.2 | 0.693 ± 0.137 |
| Senior (>=60) | 360 | 79.2 ± 7.7 | 0.655 ± 0.141 |
| *Race* | | | |
| Asian | 35 | 78.6 ± 7.9 | 0.648 ± 0.135 |
| Black | 64 | 79.0 ± 8.3 | 0.659 ± 0.145 |
| White | 639 | 80.7 ± 8.1 | 0.684 ± 0.141 |
| Others | 111 | 80.1 ± 8.3 | 0.683 ± 0.135 |

**Table 2. Performance across demographics. (A)** Our sleep staging model's accuracy gap between subgroups that differ in age, sex, and race in various datasets. Mean and standard deviation are calculated for the pairwise performance gaps for the subgroups defined by each demographic. **(B)** Performance of the sleep staging model on the MGH wireless dataset for different demographics.

**Evaluation of Predicted Respiratory Event**

Respiratory events are broadly categorized into two types: hypopnea and apnea [38]. Hypopneas involve shallow breathing, while apneas are characterized by a pause in breathing for at least 10 seconds. In our paper, the apnea and hypopnea labels follow the Centers for Medicare & Medicaid Services (CMMS) rules. The severity of such respiratory events is measured via the Apnea-Hypopnea Index (AHI) defined as the average number of apnea or hypopnea events per hour. Based on the AASM guidelines, there are 4 general categories of sleep apnea severity: Normal: AHI ≤ 5; Mild: 5 < AHI ≤ 15; Moderate: 15 < AHI ≤ 30; Severe: AHI > 30 [38].

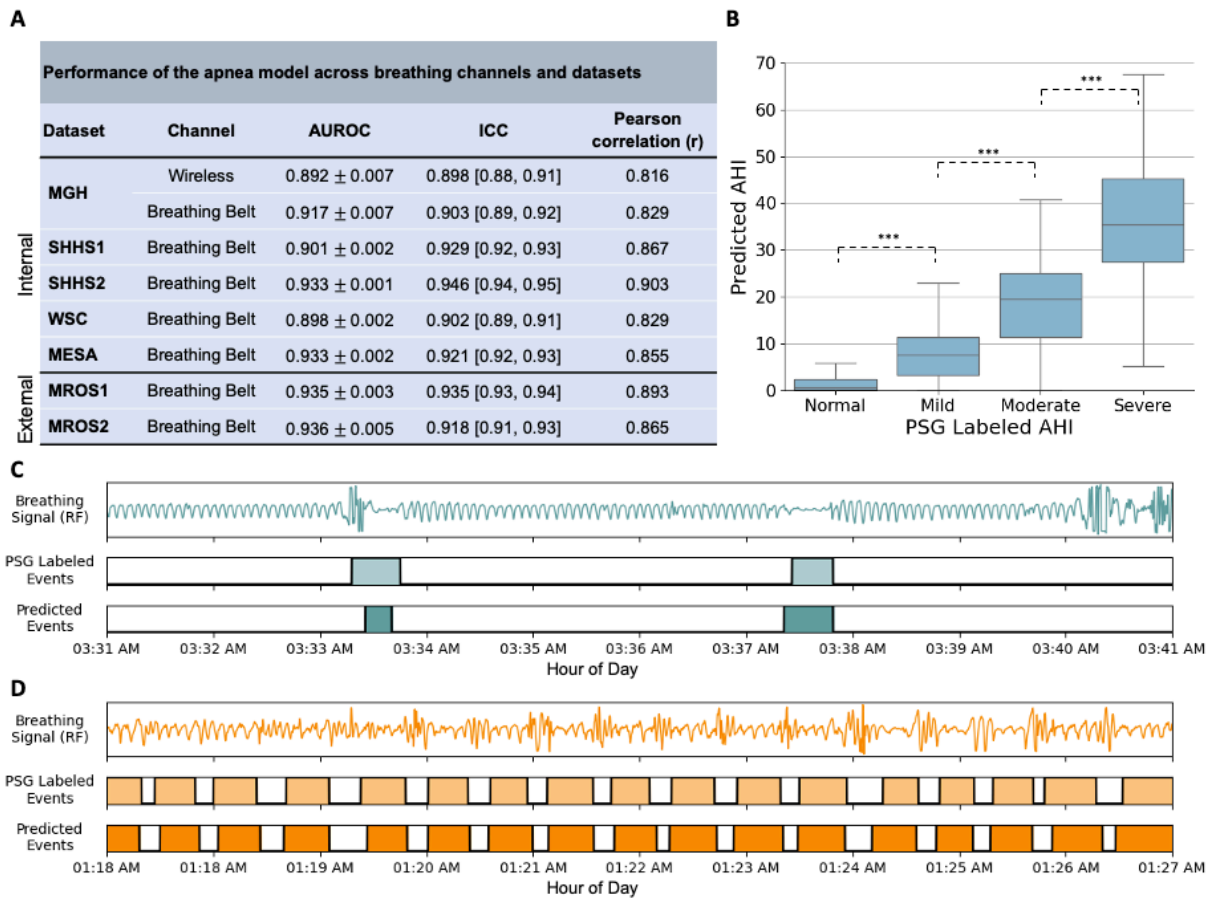

**Figure 3. Performance of the apnea model. (A)** Performance across different channels and datasets. **(B)** Comparison of predicted AHI from wireless signals (in MGH wireless dataset) across the 4 severity categories of sleep apnea. **(C-D)** Examples of the PSG labeled respiratory events with the predicted respiratory events for two participants associated with different sleep apnea severity ("Normal" **(C)** and "Severe" **(D)**).

Here, we validate that our model can detect respiratory events from breathing signals. The results, in Figure 3A, show that the model accurately assesses AHI independently whether the breathing is collected using a breathing belt or the radio signals reflected off the person's body. For example, for the MGH dataset, the intraclass correlation coefficient (ICC) for predicting the AHI is 0.90 (95% CI=[0.88,0.91]) from radio signals and 0.90 (95% CI=[0.89,0.92]) from a breathing belt (p < 0.001). A similar high accuracy is observed for the other datasets as listed in Figure 3. The performance difference between the AI model and the ground truth is similar to the performance difference between two sleep technicians who label the same data (in which case the ICC is also 0.95 [46]). We also note that the model achieves such high accuracy even on external datasets (i.e., MROS1 and MROS2) not included in its training.

We also evaluate the clinical utility of using radio signals to assess the severity of someone's sleep apnea. Figure 3B shows that the model can assess the severity of a person's sleep apnea using solely radio signals (p < 0.001). The sensitivity and specificity for AHI > 5 are 0.74 and 0.89 respectively, and the sensitivity and specificity for AHI > 15 are 0.73 and 0.95 respectively. It is worth noting that the model occasionally misclassifies participants between the Normal and Mild categories more frequently than other categories, which is reflected in the overlap in the boxes between the two categories. This can be attributed to the relatively narrower distinction in AHI values within these less severe categories. Furthermore, it is important to consider that the recommended treatment approaches for both Normal and Mild patients often overlap, as CPAP treatment is recommended only for patients with moderate to severe sleep apnea [47].

Finally, in Figure. 3C-D, we present examples of PSG-labeled respiratory events and the corresponding predictions made by our model for two patients with varying levels of sleep apnea severity (normal vs. severe). Our model successfully detects the occurrence of respiratory events in both instances, accurately capturing the presence of these events across varying degrees of sleep apnea severity.

**Robustness to Pre-existing Conditions**

A practical sleep staging model must perform well not just for healthy individuals, but also for people with pre-existing health conditions. We leverage that some of the datasets in our study have annotations of existing morbidities to assess our model's performance in four principal disease categories: cardiovascular, respiratory, immune, and neurological diseases. Note that the morbidities are not mutually exclusive, patients can have multiple diseases at the same time. Figure 4A shows that the model has good performance across all these categories for various datasets. Specifically, the average accuracy is 80%, 79%, 78%, and 78% for individuals with cardiovascular, respiratory, immune, and neurological diseases respectively. We make two additional observations related to Figure 4A. First, although our model relies on respiratory signals as input, its accuracy is not hampered by the presence of respiratory diseases. Second, the largest performance gap is associated with neurological conditions, which could be attributed to the interaction between neurological disorders and EEG, which is used to label sleep stages [2].

For additional insight, and to emphasize the ability of the model to work purely based on radio signals, we zoom in on different diseases in each of the above disease categories for the MGH wireless dataset. Figure 4B shows that the model demonstrates consistently high performance across all fine-grained health conditions in the disease categories.

**Changes in Sleep Architecture with Diseases and Drugs**

Sleep architecture is affected by diseases and drugs. In particular, REM sleep is affected by cardiovascular [48, 49], autoimmune [50, 51], and mental health [52]. In this section, we aim to validate that the sleep metrics derived from our model can faithfully capture changes in sleep architecture between cohorts who have different health conditions. We focus on REM architecture and consider four common diseases: Depression, Hypertension, Rheumatoid Arthritis, and Diabetes. For each disease, we study multiple datasets to confirm the association between the observed change in REM and the underlying condition. We also leverage that some of our datasets contain the subject's medication record to investigate whether the impact on REM sleep is due to the disease itself or the pharmaceutical intervention.

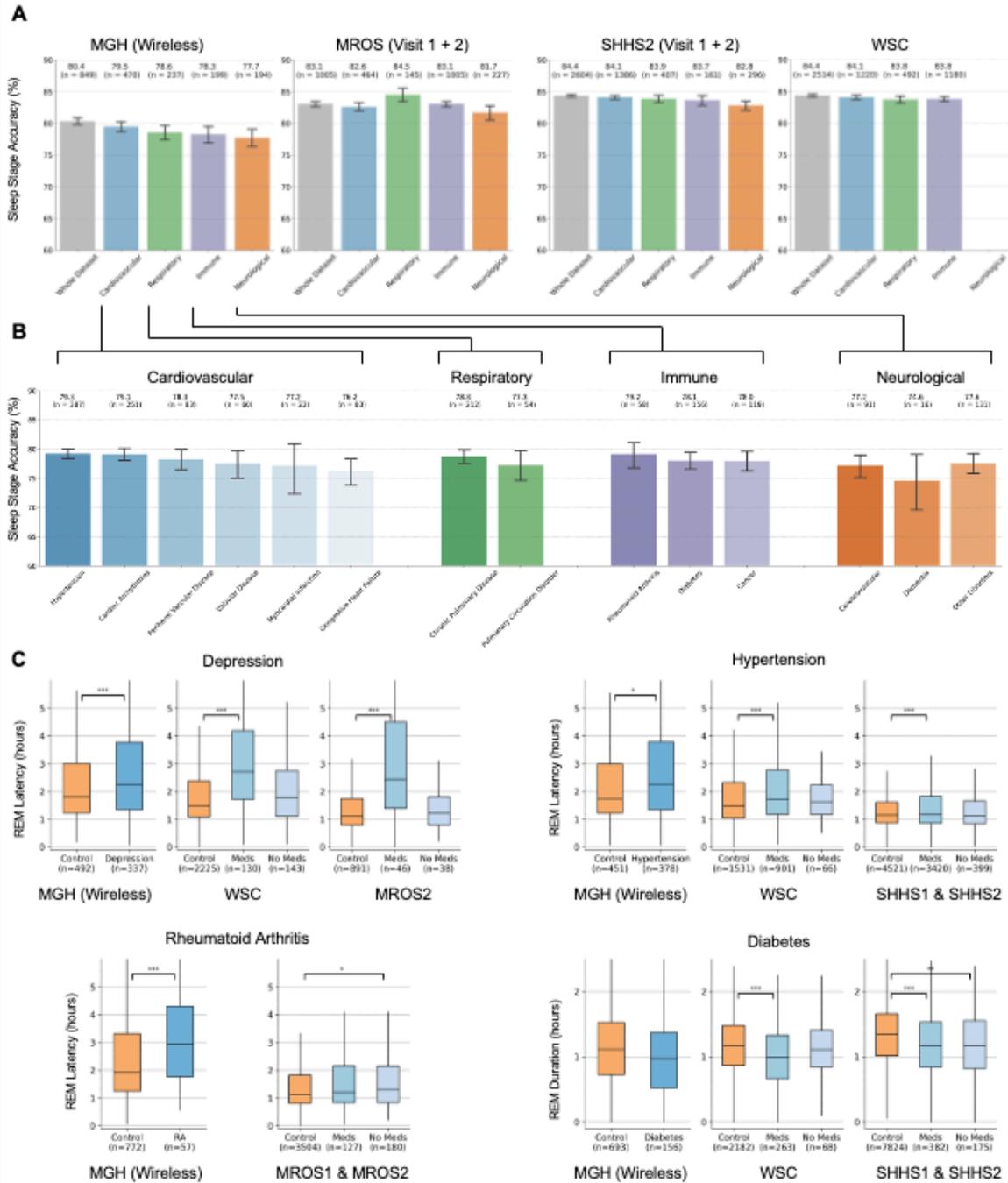

**Figure 4. Analysis of the sleep staging model in diseased population. (A)** Accuracy of the model across different disease categories (Respiratory, Cardiovascular, Immune, and Neurological diseases) in different datasets. 'All' refers to all participants in the dataset. Note that the disease categories are not mutually exclusive. **(B)** Accuracy of sleep stage prediction across fine-grained conditions in the MGH dataset using wireless signals as input. **(C)** Disease and drugs' effects on predicted sleep metrics. We consider Depression, Hypertension, Rheumatoid Arthritis (RA), and Diabetes. For each disease, we analyze the sleep metrics that have differences between the control and the disease groups. For datasets with medication information, we further split the disease group into patients on medications (Meds) and not on medications (No Meds) to study the effect of drugs. The statistical significances are indicated by asterisks. (*: p<0.05, **: p<0.01, ***: p < 0.001). Note that, compared to other datasets, patients in the MGH dataset have higher REM latency and less REM duration, which is due to patients visiting the MGH sleep lab are more likely to have sleep problems.

We report the results in Figure 4C. The REM metrics in Figure 4C are based on the output of our model. Figure S5 in the Supplementary Material provides similar graphs for the case where the REM metrics are computed from the ground truth sleep labels, demonstrating that findings made based on the AI model match those from the ground truth data. The statistical significance is computed using multiple linear regression to adjust for age and sex.

Figure 4C shows that individuals suffering from depression tend to experience higher REM latency than those who do not suffer from such a condition. However, the reason for increased REM in these cases is likely due to medications (i.e., antidepressants) as opposed to the condition itself. Similarly, higher REM latency is observed in individuals with hypertension, and the empirical evidence reveals that this latency is due to antihypertension drugs. These results are compatible with the literature which reports that antidepressants that increase serotonin function have a significant effect on REM sleep [53], and anti-hypertension drugs such as β-adrenoceptor and α2-adrenoceptor agonists reduce the duration of REM sleep [54].

The figure also shows that Rheumatoid Arthritis (RA) is associated with increased REM latency. Yet, unlike hypertension and depression, REM delay seems to stem from the disease as opposed to RA medications. Furthermore, the figure shows that diabetes is associated with less REM sleep. Here too, the data indicates that reduction in REM sleep is due to the underlying disease as opposed to its medications. Our results on the impact of RA and diabetes on REM sleep are consistent with medical literature [37-39].

## Discussion

We have demonstrated the ability to extract sleep stages and respiratory events from nocturnal breathing, measured using radio waves or a breathing belt. Our findings have been rigorously validated against the PSG gold standard (Figure 2), revealing a highly accurate sleep staging model that is equitable across various patient demographics (Figure 3) and robust against different diseases and health conditions (Figure 4). Given that our system enables the noninvasive, contactless, and passive measurement of breathing signals through radio signals, it offers a convenient means to continuously monitor sleep stages and respiratory events from the comfort of one's own home, every night.

This capability (i.e., passive and daily assessment of sleep and nocturnal respiration at home) brings several benefits to both clinical care and clinical trials. There is a strong interest in conducting home-based sleep studies for numerous reasons. Traditional sleep lab tests often burden patients with an excessive number of sensors and force them to sleep in an unfamiliar environment, both of which can disrupt sleep and bias the exact phenomenon that the test measures. Moreover, in-lab tests are costly, averaging around $3,000 per night at the time of writing. Additionally, accessibility remains an issue, particularly in rural areas and on a global scale, where hundreds of millions suffer from sleep disorders, yet PSG labs and qualified sleep technicians are scarce in many regions [55]. Perhaps most importantly, home sleep tests alleviate the burden on patients.

All these factors have spurred interest in at-home sleep tests. However, existing at-home testing kits [56, 57] are typically limited to assessing apnea and respiratory events, as they do not generate a hypnogram [56]. Furthermore, they can be cumbersome for patients, as they require the use of respiratory sensors during sleep. As a result, at-home sleep testing falls short when compared to in-lab testing.

Our results have the potential to significantly expand the scope of at-home sleep testing. On one hand, the ability to accurately extract a sleep hypnogram from respiratory signals means that typical home sleep testing devices can offer more than just apnea testing by including sleep staging. On the other hand, the fact that an equally accurate assessment can be obtained passively through radio signals without the need for wearable sensors allows for continuous monitoring of sleep and respiratory events without any inconvenience for months or even years. This greatly facilitates the conduct of longitudinal at-home sleep studies in clinical trials and the management of mood disorders and other neurological conditions where sleep problems are major contributing factors to the disease burden.

Additionally, our results have the potential to enhance access and equity in healthcare. As mentioned earlier, many communities lack access to nearby PSG labs, and some countries lack sleep labs altogether. The ability to remotely and automatically assess sleep and respiratory events empowers these communities, ensuring more equitable access to healthcare services. Access also improves for children and individuals with dementia or cognitive impairment, for whom wearing sensors can be challenging even when conducting sleep studies at home.

Furthermore, our study underscores the relationship between breathing, sleep, and EEG. We used breathing as an intermediate signal to extract a comprehensive sleep hypnogram with corresponding sleep stages. Typically, sleep staging relies on analyzing the EEG signal. Previous research has reported an association between specific respiratory patterns and sleep stages [58, 59], and the feasibility of sleep staging based on breathing [27, 60]. Our results reinforce and expand upon this body of literature by demonstrating that the relationship between breathing and sleep staging remains consistent across demographics and various health conditions, including those that directly affect respiration, such as COPD and bronchitis.

However, it is essential to acknowledge the limitations of our study. First, a PSG test provides additional data beyond the hypnogram and respiratory events (e.g., ECG, EEG, EMG). While these additional signals are helpful in certain diagnoses, sleep tests typically focus on the hypnogram and respiratory events. Second, identifying the N1 sleep stage poses a challenge for the machine learning model. N1 is characterized by its brief and transitional nature, making accurate labeling difficult even for experienced sleep technicians [30]. Some literature suggests that the labels associated with the N1 stage are inherently unreliable and offer limited clinical utility in practice [61]. Third, our evaluation did not include predicting distinct categories of respiratory events, such as hypopnea and apnea. Inter-rater agreement for these categories is notably weak compared to overall respiratory events [56], primarily due to the subjective nature of distinguishing hypopnea and apnea, introducing greater variability. Finally, our validation is in comparison to PSG lab tests, where participants naturally sleep alone. Home environments are likely more complex and people may sleep with a bed partner. While this paper does not evaluate cases where people sleep at home or have a bed partner, prior work has demonstrated that passive collection of breathing signals from radio waves remains accurate in home environments, whether people sleep alone or with a bed partner [22]. Since our AI models rely solely on breathing signals as input, their accuracy depends solely on the quality of these breathing data, which remains high in home environments.

Our research aligns with the growing interest in harnessing artificial intelligence to gain a deeper understanding of physiological signals and provide clinicians with valuable clinical insights beyond traditional lab tests. This approach holds enormous potential for offering clinicians comprehensive, longitudinal, ambulatory sleep assessments, paving the way for personalized and data-driven healthcare.

## Data Availability

The SHHS, WSC, MROS and MESA datasets are publicly available from the National Sleep Research Resource (https://sleepdata.org/datasets). Additional data that support the findings of this study are available from the corresponding author, upon reasonable request.

## Code Availability

We intend to make the code available for research purposes under a code use agreement.

**List of Figures**

**Figure 1. Illustration of traditional and RF-based sleep assessment. (A)** PSG sensors during a sleep study. **(B)** Illustration of the output of sleep studies showing an example sleep hypnogram with 5 distinct sleep stages**,** and examples of respiratory signals with hypopnea events marked in cyan, and apnea events marked in yellow. **(D)** Illustration of our RF-based sleep and apnea monitoring system. Our sensor operates like a low power radar. It transmits wireless signals that are 1/1000 the power of a home Wi-Fi router and collects their reflections from nearby people. The breathing signal is extracted from such reflections, and the sleep hypnogram and respiratory events are then predicted from each night of the nocturnal breathing signal.

**Figure 2. Performance of our sleep staging model. (A)** Accuracy of sleep staging model across breathing channels and datasets. **(B)** Agreement of model's prediction from different breathing channels, evaluated by Cohen's Kappa. Each entry shows the mean and 95% confidence interval. **(C)** Confusion matrices of predicted sleep stages from radio waves in the MGH dataset. Left is 4 class staging; Right is 5 class staging. **(D)** Example of a ground truth hypnogram from PSG data and the corresponding hypnogram output by our model based on radio signals. The 4 sleep cycles during the night are separated by the dashed lines. **(E)** Scatter plots of predicted sleep metrics from radio signals versus the ground truth from PSG data. Metrics are Total Sleep Time (TST), Sleep Efficiency (SE), Sleep Onset Latency (SOL), and Wake After Sleep Onset (WASO). Pearson correlations and p-values from the two-tailed Wald Test with t-distribution are reported.

**Figure 3. Performance of the apnea model. (A)** Performance across different channels and datasets. **(B)** Comparison of predicted AHI from wireless signals (in MGH wireless dataset) across the 4 severity categories of sleep apnea. **(C-D)** Examples of the PSG labeled respiratory events with the predicted respiratory events for two participants associated with different sleep apnea severity ("Normal" **(C)** and "Severe" **(D)**)

**Figure 4. Analysis of the sleep staging model in diseased population. (A)** Accuracy of the model across different disease categories (Respiratory, Cardiovascular, Immune, and Neurological diseases) in different datasets. 'All' refers to all participants in the dataset. Note that the disease categories are not mutually exclusive. **(B)** Accuracy of sleep stage prediction across fine-grained conditions in the MGH dataset using wireless signals as input. **(C)** Disease and drugs' effects on predicted sleep metrics. We consider Depression, Hypertension, Rheumatoid Arthritis (RA), and Diabetes. For each disease, we analyze the sleep metrics that have differences between the control and the disease groups. For datasets with medication information, we further split the disease group into patients on medications (Meds) and not on medications (No Meds) to study the effect of drugs. The statistical significances are indicated by asterisks. (*: $p<0.05$, **: $p<0.01$, ***: $p < 0.001$). Note that, compared to other datasets, patients in the MGH dataset have higher REM latency and less REM duration, which is due to patients visiting the MGH sleep lab are more likely to have sleep problems.

## Acknowledgment

We would like to express our gratitude to the members of the Katabi's lab at MIT for their comments on our manuscript. We also thank Professor Wenlong Mou from the University of Toronto for his thorough review of the statistical methods in the paper and his insightful feedback. Additionally we extend our thanks to all the research participants for their time and contributions. The content of this paper is solely the responsibility of the authors and does not necessarily represent the official views of NSF or other sponsors.

## Funding

This work is funded by the National Science Foundation Award No. IIS-2014391.

## Financial Disclosure

D.K. receives research funding from the NIH, NSF, Sanofi, Takada, IBM, Gwangju Institute of Science and Technology, Michael J Fox Foundation, Helmsley Charitable Trust, and the Rett Syndrome Research Trust. She is a co-founder of Emerald Innovations and has a personal equity interest in the company. She also serves on the board of directors of Cyclerion, the scientific advisory board of Janssen, and the data science advisory board of Amgen. B.W. receives funding from the NIH and NSF. He is a co-founder, scientific advisor, and consultant to Beacon Biosignals and has a personal equity interest in the company. J.S. receives research funding from the NSF. These relationships played no role in the present study. The rest of the authors have no financial disclosures. R.H. is a co-founder of Emerald Innovations and has a personal equity interest in the company.

## Non-financial Disclosure

None

## Author Contributions

D.K. conceived the idea of contactless PSG from radio waves. H.H., C.L., and D.K. developed the machine learning models and algorithms. D.K., J.S. and B.W. conceived the collaboration and secured the funding. B.W. provided clinical insights. R.H. deployed the radio devices in the sleep lab. W.G., K. G., H.S., and B.W. provided the anonymized MGH Sleep Lab dataset and the corresponding demographic and comorbidities information. H.H. and C.L. processed the data and performed experimental validation. H.H., C.L., M.O., and D.K. conducted the analysis. H.H. and C.L. generated the figures. H.H., C.L., and D.K. wrote the original manuscript. D.K. supervised the work. All authors reviewed and approved the manuscript.